\documentclass[11pt,a4paper]{article}
\usepackage{times}
\usepackage{latexsym}
\usepackage{booktabs} 
\usepackage{wrapfig}
\usepackage{xspace,amsmath,amsfonts,array,graphicx,subcaption}
\usepackage{algorithm}
\usepackage{algpseudocode}
\usepackage{pgfplots}
\usepackage{microtype}
\usepackage{pifont}
\usepackage{color, colortbl}
\usepackage{float}
\usepackage{dblfloatfix}
\usepackage[hyphens]{url} %
\usepackage[nohyperref]{naaclhlt2018}
\usepackage{siunitx}
\usepackage{caption}

\aclfinalcopy

\newif\iffinal
\finalfalse
\finaltrue

\newcommand{\ftext}{$f_{\text{text}}$\xspace}
\newcommand{\fimage}{$f_{\text{image}}$\xspace}
\newcommand{\mathfimage}{f_{\text{image}}}
\newcommand{\mathftext}{f_{\text{text}}}

\newcommand{\numwikiarticle}{192K\xspace}
\newcommand{\numwikiimages}{549K\xspace}

\newcommand{\mparagraph}[1]{\smallskip \noindent\textbf{{#1}}.}
\newcommand{\mparagraphnp}[1]{\noindent\textbf{{#1}}}

\newcommand{\concrete}{concrete\xspace}
\newcommand{\concreteness}{concreteness\xspace}

\newcommand{\Concreteness}{Concreteness\xspace}

\newcommand{\mni}{{\rm MNI}^k\xspace}
\newcommand{\knn}{{\rm NN}^k\xspace}

\newcommand{\rability}{retrievability\xspace}
\newcommand{\Rability}{Retrievability\xspace}

\definecolor{firstplacecolor}{rgb}{0.2980392156862745,0.4470588235294118,0.6901960784313725}
\definecolor{secondplacecolor}{rgb}{0.3333333333333333,0.6588235294117647,0.40784313725490196}
\definecolor{thirdplacecolor}{rgb}{0.7686274509803922,0.3058823529411765,0.3215686274509804}

\newcommand{\firstplace}[1]{\textbf{\textcolor{firstplacecolor}{#1}}}
\newcommand{\secondplace}[1]{\textbf{\textcolor{secondplacecolor}{#1}}}
\newcommand{\thirdplace}[1]{\textbf{\textcolor{thirdplacecolor}{#1}}}
\title{
Quantifying the visual \concreteness of words and topics\\ in multimodal datasets}

\author{
  Jack Hessel \\
  Cornell University \\
  {\tt jhessel@cs.cornell.edu} \\\And
  David Mimno \\
  Cornell University \\
  {\tt mimno@cornell.edu} \\\And
  Lillian Lee \\
  Cornell University \\
  {\tt llee@cs.cornell.edu}
}

\date{}
\begin{document}
\maketitle
\begin{abstract}
  Multimodal machine learning algorithms aim to learn visual-textual
  correspondences. Previous work suggests that concepts with
  \emph{concrete} visual manifestations may be easier to learn than
  concepts with abstract ones. We give an algorithm for automatically
  computing the visual concreteness of words and topics within
  multimodal datasets. We apply the approach in four settings, ranging
  from image captions to images/text scraped from historical books. In
  addition to enabling explorations of concepts in multimodal
  datasets, our concreteness scores predict the capacity of machine
  learning algorithms to learn textual/visual relationships. We find
  that 1) concrete concepts are indeed easier to learn; 2) the large
  number of algorithms we consider have similar failure cases; 3) the
  precise positive relationship between concreteness and performance
  varies between datasets. We conclude with recommendations for using
  concreteness scores to facilitate future multimodal research.
\end{abstract}


\section{Introduction}

Text and images are often used
together
to
serve as a richer form of content. For example, news articles may be accompanied by
photographs or infographics; images shared on social media are often coupled
with descriptions or tags; and
textbooks
include illustrations, photos, and other visual elements. The ubiquity and diversity of
such ``text+image''  material (henceforth referred to as \emph{multimodal} content)
 suggest that, from the standpoint of sharing
information, images and text are
often
natural complements.

\begin{figure*}[t]
\centering
\includegraphics[width=.95\linewidth]{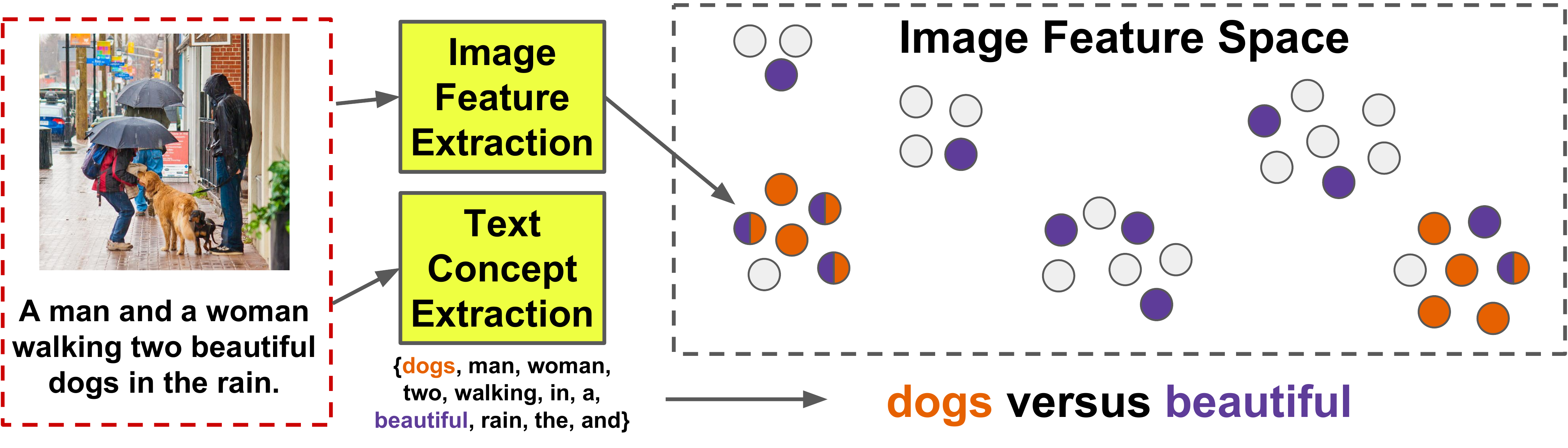}
    \caption{Demonstration of visual \concreteness estimation
      on an
      example from the COCO dataset.
      The degree of visual clustering of
      textual concepts is measured using a nearest neighbor
      technique. The \concreteness of ``dogs'' is greater than the
      \concreteness of ``beautiful'' because images associated with
      ``dogs'' are packed tightly into two clusters, while images
      associated with ``beautiful'' are spread evenly.\protect\footnotemark}
    \label{fig:cohesiveness}
\end{figure*}

Ideally, machine learning algorithms that incorporate information from
both text and images should have a fuller perspective than those that
consider either text or images in isolation.
But \newcite{hill2014learning} observe that for their particular
multimodal architecture, the level of {\em \concreteness} of a concept
being represented ---
intuitively, the idea of a {\em dog} is more concrete than that of
{\em beauty} --- affects whether multimodal or single-channel
representations are more effective.  In their case, \concreteness was
derived for 766 nouns and verbs from a fixed psycholinguistic database
of human ratings.

In contrast, we introduce an adaptive algorithm for characterizing the
visual
\concreteness of
all the concepts indexed textually (e.g., ``dog'') in a given multimodal dataset.
Our approach is to
leverage the geometry of image/text space.
Intuitively,
a visually concrete concept is one associated with more
locally similar sets of
images; for example, images associated with ``dog'' will likely
contain dogs, whereas images associated with ``beautiful''  may contain
flowers, sunsets,
weddings, or an abundance of other possibilities --- see Fig.~\ref{fig:cohesiveness}.

  Allowing \concreteness to be
dataset-specific is an important innovation because \concreteness is
contextual.  For example, in one dataset we work with, our method
scores ``London'' as highly \concrete because of a preponderance of
iconic London images in
it,
such as Big Ben and double-decker
buses; whereas for a separate dataset, ``London'' is used as a geotag
for diverse images, so the same word scores as highly non-\concrete.

In addition to being dataset-specific, our method readily scales, does
not depend on an external search engine, and is compatible with both discrete and
continuous textual concepts (e.g., topic distributions).

Dataset-specific visual \concreteness scores
enable
a variety of purposes. In this paper,
we focus on using them to: 1) explore multimodal datasets; and 2)
predict how easily concepts will be learned in a machine learning
setting.
\footnotetext{Image copyright information is provided in the supplementary material.}
We apply our method to four large multimodal datasets, ranging from
image captions to image/text data scraped from
Wikipedia,\footnote{
We release our Wikipedia and British Library data at \url{http://www.cs.cornell.edu/~jhessel/concreteness/concreteness.html} }
to
  examine the relationship between
\concreteness scores and the performance of machine learning
algorithms. Specifically, we consider the cross-modal retrieval problem, and examine
a
number of NLP, vision, and retrieval algorithms. Across all
320  significantly different experimental settings (=
4 datasets
$\times$ 2 image-representation algorithms $\times$ 5 textual-representation algorithms $\times$ 4 text/image alignment
algorithms
$\times$ 2 feature pre-processing schemes),
we find that more
\concrete instances are easier to retrieve, and that different algorithms have
similar failure cases. Interestingly, the relationship between
\concreteness and retrievability varies significantly based on
dataset: some datasets appear to have a linear relationship
between the two, whereas others exhibit a \concreteness threshold beyond
which retrieval becomes much easier.

We believe that our work can have a positive impact on future multimodal research.
\S \ref{sec:future} gives more detail, but in brief, we see implications in (1) evaluation
--- more credit should
perhaps be assigned to performance on non-\concrete concepts; (2) creating or augmenting multimodal datasets, where
one might {\em a priori} consider the desired relative proportion of \concrete vs. non-\concrete concepts; and (3) {\em
curriculum learning}
\cite{bengio2009curriculum}, where ordering of training examples could take \concreteness levels into account.

\section{Related Work}

Applying machine learning to understand visual-textual relationships
has enabled a number of new applications, e.g., better accessibility
via
automatic generation of alt text
\citep{Garcia+Paluri+Wu:2016a},
cheaper
training-data acquisition for computer vision
\cite{joulin2016learning,veit2017learning}, and cross-modal retrieval
systems, e.g.,
\citet{rasiwasia2010new,costa2014role}.

Multimodal datasets often have substantially differing
characteristics, and are used for different tasks
\citep{baltruvsaitis2017multimodal}.
Some commonly used
datasets couple images with a handful of unordered tags
\cite[inter alia]{barnard2003matching,cusano2003image,grangier2008discriminative,chen2013fast}
or
short, literal natural language captions
\cite[inter alia]{farhadi2010every,ordonez2011im2text,kulkarni2013babytalk,fang2015captions}.
In other cross-modal retrieval settings, images are paired with long, only loosely thematically-related documents.
\citep[inter alia]{khan2009tvgraz,socher2010connecting,jia2011learning,zhuang2013supervised}. We
provide experimental results on both types of data.

\Concreteness in datasets has been previously studied in either
text-only cases \cite{turney2011literal,hill2013concreteness} or by
incorporating human judgments of perception into models
\cite{silberer2012grounded,hill2014concreteness}. Other work has
quantified characteristics of concreteness in multimodal datasets
\cite{young2014image,hill2014multi,hill2014learning,kiela2014learning,jas2015image,lazaridou2015combining,silberer2016visually,lu2016knowing,bhaskar2017exploring}.
Most related to
our work is that of Kiela et al. \shortcite{kiela2014improving}; the
authors use Google image search to collect 50 images each for a
variety of words and compute the average cosine similarity between
vector representations of returned images. In contrast, our method can
be tuned to specific datasets
without reliance on an external search engine.
Other algorithmic advantages of our method include that: it
more readily scales than previous solutions, it makes relatively few
assumptions regarding the distribution of images/text, it normalizes
for word frequency in a principled fashion, and it can produce
confidence intervals. Finally, the method we propose
can be applied to
both
discrete and continuous concepts like topic distributions.


\section{Quantifying Visual \Concreteness} \label{quantconc}

 To compute visual
\concreteness scores, we
adopt the same general approach as
\newcite{kiela2014improving}:
for a fixed
text concept (i.e., a
word or topic), we measure the
variance in the corresponding
visual
features. The method is summarized in
Figure~\ref{fig:cohesiveness}.

\subsection{\Concreteness of discrete words}

We assume as input
a multimodal dataset
of $n$ images represented
in a space where nearest neighbors may be computed.
Additionally, each
image is associated with a set of discrete words/tags. We write $w_v$
for the set of words/tags associated with image $v$, and $V_w$ for the
set of all images associated with a word $w$. For example, if the
$v^{th}$ image is of a dog playing frisbee, $w_v$ might be
$\{\text{frisbee}, \text{dog}, \text{in}, \text{park}\}$, and $v \in
V_{\text{park}}$.

Our goal is to measure how ``clustered'' a word is in image feature
space.
Specifically, we ask: for each image $v \in V_w$,
how often are $v$'s
nearest neighbors also associated with $w$? We
thus compute the expected value of {\em $\mni_w$}, the number of \underline{m}utually \underline{n}eighboring \underline{i}mages of word $w$:
\begin{equation}
\mathbb{E}_{P_{data}}[\mni_w] = \frac{1}{|V_w|} \sum_{v \in V_w} | \knn(v) \cap V_w | \, ,
\label{eq:empirical_mni}
\end{equation}
where
$\knn(v)$ denotes the set of $v$'s $k$ nearest
neighbors in image space.

While Equation~\ref{eq:empirical_mni} measures clusteredness, it does
not properly normalize for frequency. Consider a word like ``and''; we
expect
it
to have low \concreteness, but
its
associated
images will share neighbors simply because ``and'' is a frequent
unigram. To correct for this, we compute the {\em \concreteness} of a
word as the ratio of $\mathbb{E}[\mni_w]$ under the true distribution
of the image data to a random distribution of the image data:
\begin{equation}
\text{\concreteness}(w) = \frac{\mathbb{E}_{P_{data}}[\mni_w]}{\mathbb{E}_{P_{random}}[\mni_w]}
\label{eq:cohesiveness}
\end{equation}
While the denominator of this expression can be computed in closed form,
we use $\mathbb{E}_{P_{random}}[\mni_w] \approx
\frac{k|V_w|}{n}$; this approximation is faster to compute and is
negligibly different from the true expectation in practice.

\subsection{Extension to continuous topics}

We extend the definition of \concreteness to continuous concepts, so
that our work applies also to topic model outputs; this extension is
needed because the intersection
in Equation~\ref{eq:empirical_mni}
cannot be directly applied to real values. Assume we are given
a set of topics $T$ and an
image-by-topic matrix
$Y \in \mathbb{R}^{n\times |T|}$, where the $v^{th}$ row\footnote{
  The construction is necessarily different for different types of datasets,
  as described in \S \ref{sec:data}.
  }
is a topic distribution
for the text associated with image $v$, i.e.,
$Y_{ij} = P(\text{topic } j | \text{image } i)$. For each topic $t$,
we compute the average topic weight for each image $v$'s neighbors,
and take a weighted average as:
\begin{equation}
  \text{\concreteness}(t) = \frac{n}{k}\cdot \frac{\sum_{v=1}^n \lbrack Y_{vt} \sum_{j \in \knn(v)} Y_{jt} \rbrack}{\left(\sum_{v=1}^n Y_{vt}\right)^2}
  \label{eq:continuous}
\end{equation}

Note that Equations~\ref{eq:empirical_mni} and \ref{eq:continuous} are
computations of means. Therefore, confidence intervals can be computed
in both cases either using a normality assumption or bootstrapping.


\section{Datasets}
\label{sec:data}

We consider four datasets that span a variety of multimodal settings.
Two are publicly available and widely used (COCO/Flickr); we collected
and preprocessed the other two (Wiki/BL). The Wikipedia and British
Library sets are available for download at
\url{http://www.cs.cornell.edu/~jhessel/concreteness/concreteness.html}.
Dataset statistics are given in Table~\ref{tab:dataset_stats}, and
summarized as follows:

\begin{figure}
  \centering
  \includegraphics[width=.95\linewidth]{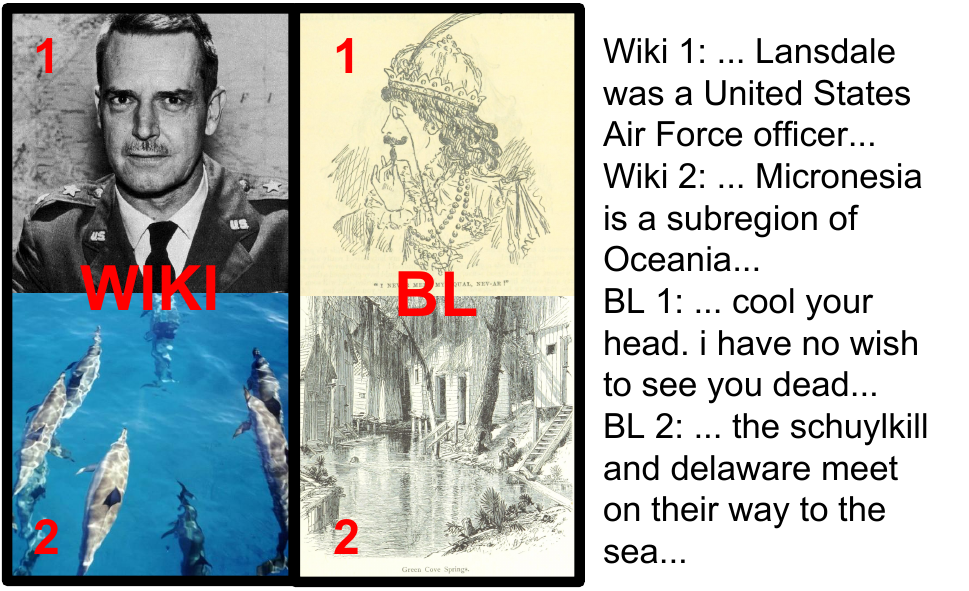}
    \caption{Examples of text and images from
    our new Wiki/BL datasets.}
    \label{fig:dataset-examples}
\end{figure}

\mparagraph{Wikipedia (Wiki)} We collected a dataset
consisting of \numwikiarticle articles from the English Wikipedia,
along with the \numwikiimages images contained in those
articles. Following Wilson's popularity filtering
technique,\footnote{\url{https://goo.gl/B11yyO}} we selected this
subset of Wikipedia by identifying articles that received at least 50
views on March 5th, 2016.\footnote{The articles were extracted from an
  early March, 2016 data dump.}
To our knowledge, the previous largest publicly available multimodal
Wikipedia dataset comes from ImageCLEF's 2011 retrieval task
\cite{popescu2010overview}, which consists of 137K images associated
with English articles.

\begin{table}
  \begin{tabular}{l|lS[table-format=4.1]r@{$/$}r}
& \# Images & {Mean Len} & Train & Test \\
\midrule
Wiki & 549K & 1397.8 & 177K & 10K \\
BL & 405K & 2269.6 & 69K & 5K \\
COCO & 123K & 10.5 & 568K & 10K \\
Flickr & 754K & 9.0 & 744K & 10K \\
\bottomrule
\end{tabular}

  \caption{
    Dataset statistics:
    total number
    of images,
    average text length in words, and size of the
    train/test splits we use in \S \ref{sec:correspondances}.
  }
  \label{tab:dataset_stats}
\end{table}

Images often appear on multiple pages: an
image of the Eiffel tower might appear on pages for Paris, for Gustave
Eiffel, and for the tower itself.

\mparagraph{Historical Books from British Library (\textbf{BL})}
 The
British Library has released a set of digitized
books \cite{British_Library_Labs_2016} consisting of 25M pages of
OCRed text, alongside 500K+ images scraped from those pages of text.
The release splits images into four categories; we ignore ``bound
covers'' and ``embellishments'' and use images identified as
``plates'' and ``medium sized.''
We associated images with all text within a 3-page window.

 This raw data collection is noisy. Many
books are not in English, some books contain far more images than
others, and the images themselves are of varying size and rotation.
To combat these issues we only keep books that have identifiably
English text; for each cross-validation split
in our machine-learning experiments (\S\ref{sec:correspondances})
we sample at most 10 images from each book;
and we use \textit{book-level} holdout so that no images/text in the test
set are from books in the training set.

\mparagraph{Captions and Tags} We also examine two
popular existing
datasets: Microsoft COCO (captions) \cite{lin2014microsoft}
(\textbf{COCO}) and MIRFLICKR-1M (tags) \cite{huiskes2010new}
(\textbf{Flickr}). For COCO, we construct our own training/validation
splits from the 123K images, each of which has 5 captions. For Flickr,
as an initial preprocessing step we only consider the 7.3K tags that
appear at least 200 times, and the 754K images that are associated
with at least 3 of the 7.3K valid tags.


\section{Validation of \Concreteness Scoring}

\begin{figure*}
  \centering
  \includegraphics[width=.98\linewidth]{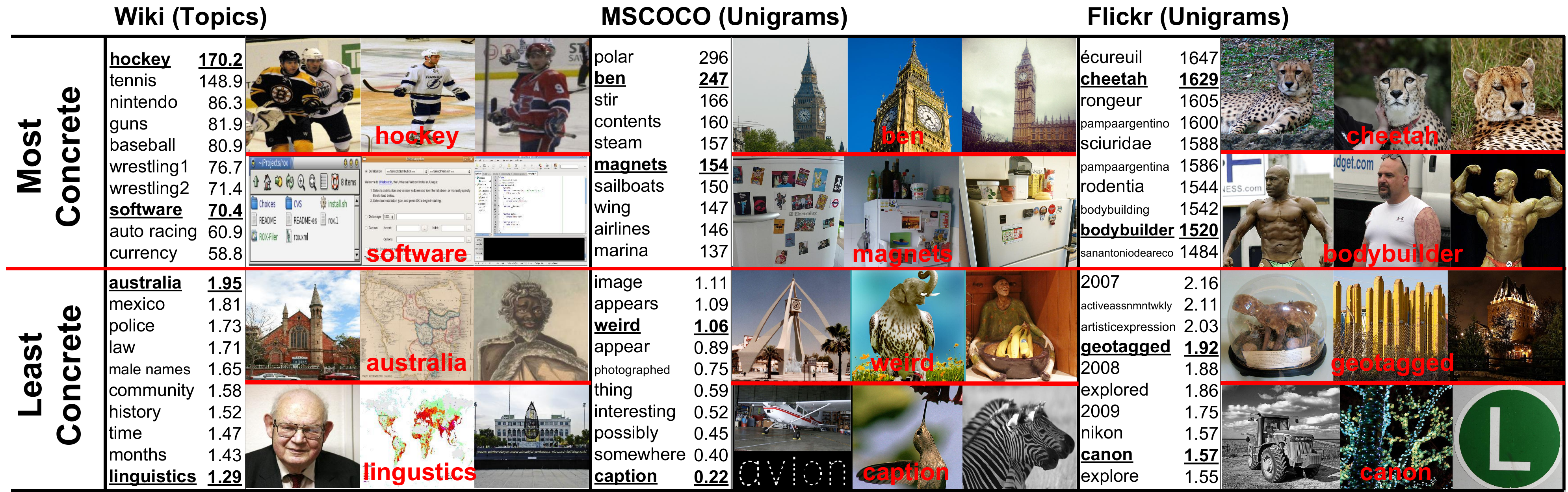}
  \caption{Examples of the most and least concrete words/topics from
    Wiki, COCO, and Flickr, along with example images associated with
    each \underline{\textbf{highlighted}} word/topic.}
  \label{fig:concrete_examples}
\end{figure*}

 We apply our \concreteness
measure to the four datasets. For COCO and Flickr, we use unigrams
as concepts,
while for Wiki and BL, we extract 256-dimensional topic distributions
using Latent Dirichlet Allocation (LDA) \cite{blei2003latent}.
For BL, topic distributions are derived from text in the
aforementioned 3 page window; for Wiki, for each image, we compute the
mean topic distribution of all articles that image appears in; for
Flickr, we associate images with all of their tags; for COCO, we
concatenate all captions for a given image. For computing
\concreteness scores for COCO/Flickr, we only consider unigrams
associated with at least 100 images, so as to ensure the stability of
MNI as defined in Equation~\ref{eq:empirical_mni}.

We extract image features from the pre-softmax layer of a deep
convolutional neural network, ResNet50
\cite{he2015deep}, pretrained
for the ImageNet classification task \cite{deng2009imagenet}; this
method is known to be a strong baseline
\cite{sharif2014cnn}.\footnote{We explore different image/text
  representations in later sections.} For nearest neighbor search, we
use the Annoy library,\footnote{\texttt{github.com/spotify/annoy}}
which computes approximate kNN efficiently. We use $k=50$ nearest
neighbors, though the results presented are stable for reasonable
choices of $k$, e.g., $k=25,100$.

\subsection{\Concreteness and human judgments} Following Kiela et
al. \shortcite{kiela2014improving}, we borrow a dataset of human
judgments to validate our \concreteness computation method.\footnote{
  Note that because \concreteness of words/topics varies from dataset
  to dataset, we don't expect one set of human judgments to correlate
  perfectly with our \concreteness scores. However, partial
  correlation with human judgment offers a common-sense ``reality
  check.''}  The concreteness of words is a topic of interest in
psychology because concreteness relates to a variety of aspects of
human behavior, e.g., language acquisition, memory,
etc. \newcite{schwanenflugel1983differential,paivio1991dual,walker1999concrete,de2000hard}.

We compare against the human-gathered unigram concreteness judgments
provided in the USF Norms dataset (USF) \cite{nelson2004university};
for each unigram, raters provided judgments of its concreteness on a
1-7 scale. For Flickr/COCO, we compute Spearman correlation using
these per-unigram scores (the vocabulary overlap between USF and
Flickr/COCO is 1.3K/1.6K), and for Wiki/BL, we compute topic-level
human judgment scores via a simple average amongst the top 100 most
probable words in the topic.

As a
 null hypothesis, we consider the possibility that our
\concreteness measure is simply mirroring frequency
information.\footnote{We return to this hypothesis in \S
  \ref{sec:performance}
as well; there, too, we find that
  \concreteness and frequency capture different information.} We
measure frequency for each dataset by measuring how often a particular
word/topic appears in it. A useful \concreteness measure should
correlate with USF more
than a simple frequency baseline
does.

\begin{figure}[h]
  \centering
  \includegraphics[width=.85\linewidth]{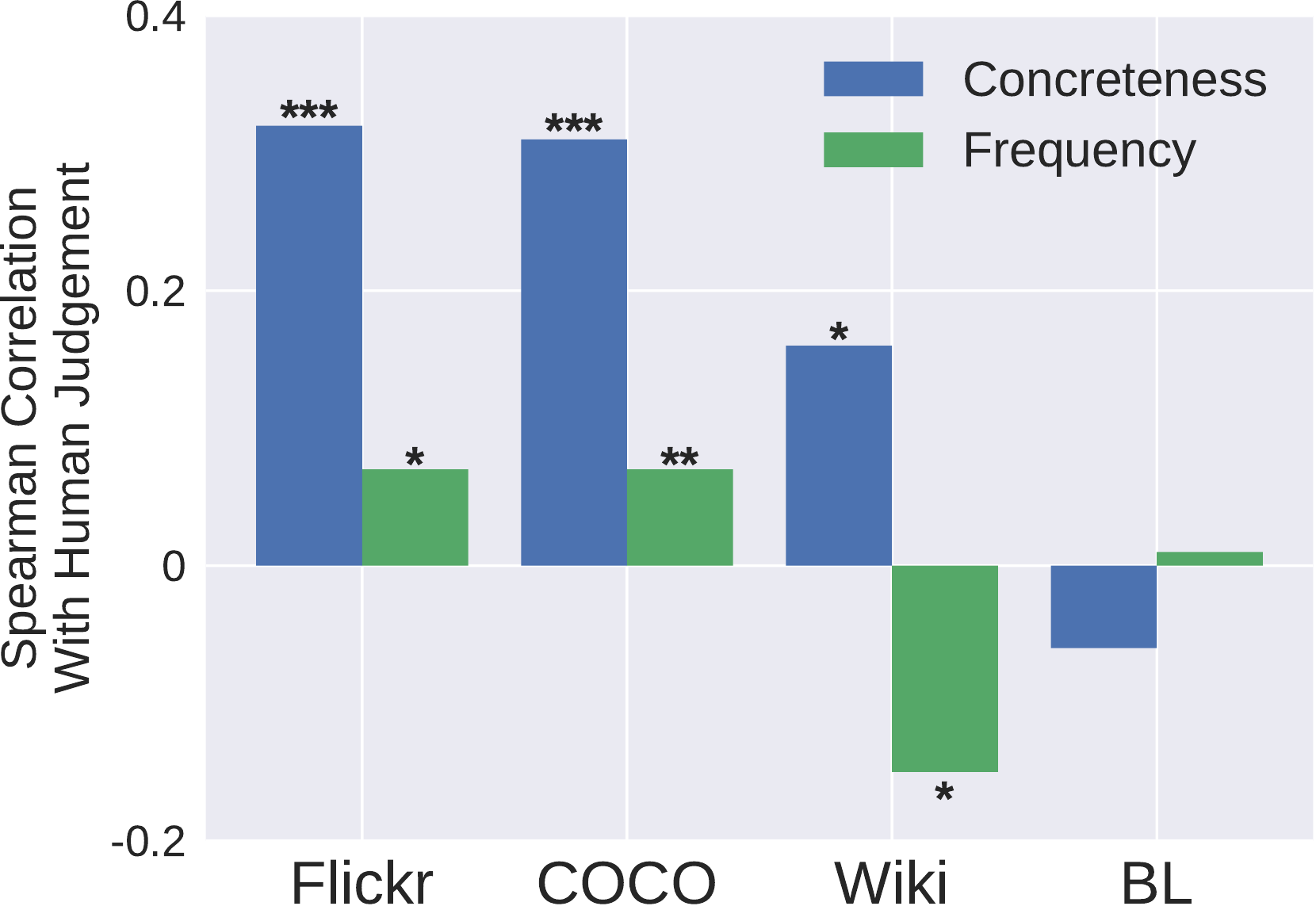}
\caption{Spearman correlations between human judgment (USF) and our
  algorithm's outputs, and dataset frequency. In the case of
  Flickr/COCO/WIKI our \concreteness scores correlate with human
  judgement to a greater extent than frequency. For BL, neither
  frequency nor our concreteness measure is correlated with human
  judgement. ***/**/* := $p < .001/.01/.05$}
\label{fig:human}
\end{figure}

For COCO/Flickr/Wiki, \concreteness scores output by our method
positively correlate with human judgments of concreteness more than
frequency does
(see Figure~\ref{fig:human}). For COCO, this pattern holds even when
controlling for part-of-speech (not shown), whereas Flickr adjectives
are not correlated with USF. For BL, neither frequency nor our
\concreteness scores are significantly correlated with USF.
Thus, in three of our four datasets, our
measure tends to predict human
concreteness judgments better than frequency.

\mparagraph{\Concreteness and frequency} While \concreteness measures
correlate with human
judgment better than frequency, we do expect
\emph{some} correlation between a word's frequency and its
\concreteness \cite{gorman1961recognition}. In all cases, we observe a
moderate-to-strong positive correlation between infrequency and
\concreteness ($\rho_{wiki},\rho_{coco},\rho_{flickr}, \rho_{bl} =
.06, .35, .40, .71$) indicating that rarer words/topics are more
concrete, in general. However, the correlation is not perfect, and
\concreteness and frequency measure different properties of words.

\subsection{\Concreteness within datasets}
Figure~\ref{fig:concrete_examples} gives examples from Wiki, COCO, and
Flickr illustrating the concepts associated with the smallest and
largest \concreteness scores according to our method.\footnote{The BL
  results are less interpretable and are omitted for space reasons.}
The scores often align with intuition, e.g., for Wiki, sports topics
are often concrete, whereas country-based or abstract-idea-based
topics are not.\footnote{Perhaps fittingly, the ``linguistics'' topic
  (top words: term, word, common, list, names, called, form, refer,
  meaning) is the least visually \concrete of all 256 topics.} For
COCO, \emph{polar} (because of polar bears) and \emph{ben} (because of
Big Ben) are concrete; whereas \emph{somewhere} and \emph{possibly}
are associated with a wide variety of images.

\Concreteness scores form a continuum, making explicit not only the
extrema (as in Figure~\ref{fig:concrete_examples}) but also the
middle ground, e.g., in COCO, ``wilderness'' (rank 479) is more
visually \concrete than ``outside'' (rank 2012).
Also,
dataset-specific intricacies that are not obvious \emph{a priori} are
highlighted, e.g., in COCO, 150/151 references to ``magnets'' (rank 6)
are in the visual context of a refrigerator (making ``magnets''
visually \concrete) though the converse is not true, as both
``refrigerator'' (rank 329) and ``fridge'' (rank 272) often appear
without magnets; 61 captions in COCO are exactly ``There is no image
here to provide a \emph{caption} for,'' and this dataset error is made
explicit through \concreteness score computations.

\subsection{\Concreteness varies across datasets}

To what extent are the \concreteness scores dataset-specific? To
investigate this question, we compute the correlation between Flickr
and COCO unigram \concreteness scores for 1129 overlapping
terms. While the two are positively correlated ($\rho=.48, p<.01$)
there are many exceptions that highlight the utility of computing
dataset-independent scores. For instance, ``London'' is extremely
concrete in COCO (rank 9) as compared to in Flickr (rank
1110).
In COCO, images of London tend to be iconic (i.e., Big
Ben, double decker buses); in contrast, ``London'' often serves as a
geotag for a wider variety of images in Flickr. Conversely, ``watch''
in Flickr is concrete (rank 196) as it tends to refer to the
timepiece, whereas ``watch'' is not concrete in COCO (rank 958) as it
tends to refer to the verb; while these relationships are not obvious
\emph{a priori}, our concreteness method has helped to highlight these
usage differences between the image tagging and captioning datasets.


\section{Learning Image/Text Correspondences} \label{sec:correspondances}
Previous work suggests that incorporating visual features for less
concrete concepts can be harmful in word similarity tasks
\cite{hill2014learning,kiela2014improving,kiela2014learning,hill2014multi}. However,
it is less clear if this intuition applies to more practical
 tasks (e.g., retrieval), or
if this problem can be overcome simply by applying the ``right''
machine learning algorithm. We aim to tackle these questions in this
section.

\mparagraph{The learning task} The task we consider is the
construction of a joint embedding of images and text into a shared
vector space. Truly corresponding image/text pairs (e.g., if the text
is a caption of that image) should be placed close together in the new
space relative to image/text pairs that do not match. This task is a
good representative of multimodal learning because computing a joint
embedding of text and images is often a ``first step'' for downstream
tasks, e.g., cross-modal retrieval \cite{rasiwasia2010new}, image
tagging \cite{chen2013fast}, and caption generation
\cite{kiros2014unifying}.

\mparagraph{Evaluations} Following previous work in cross-modal
retrieval,
 we measure performance
using the top-$k\%$ hit rate (also called recall-at-$k$-percent, $R@k\%$;
higher is better). Cross-modal retrieval can be applied in either
direction, i.e., searching for an image given a body of text, or
vice-versa. We examine both the image-search-text and
text-search-image cases
.  For simplicity, we average retrieval performance
from both directions, producing a single metric;\footnote{Averaging is
  done for ease of presentation; the performance in both directions is
  similar. Among the parametric approaches (LS/DCCA/NS) across all
  datasets/NLP algorithms, the mean difference in performance between
  the directions is 1.7\% (std. dev=2\%).} higher is better.

\mparagraph{Visual Representations} Echoing \newcite{wei2016cross}, we
find that features extracted from convolutional neural networks (CNNs)
outperform classical computer vision descriptors (e.g., color
histograms) for multimodal retrieval. We consider two different CNNs
pretrained on different datasets: ResNet50 features trained on the
ImageNet classification task (\textbf{RN-Imagenet}), and InceptionV3
\cite{szegedy2015going} trained on the OpenImages
\cite{krasin2016openimages} image tagging task
(\textbf{I3-OpenImages}).

\newcommand{\recallatfile}{01}
\newcommand{\recallat}{1}
\begin{figure*}[!t]
  \begin{subtable}{.5\linewidth}
    \begin{minipage}{.34\linewidth}
      \centering
      \includegraphics[width=\linewidth]{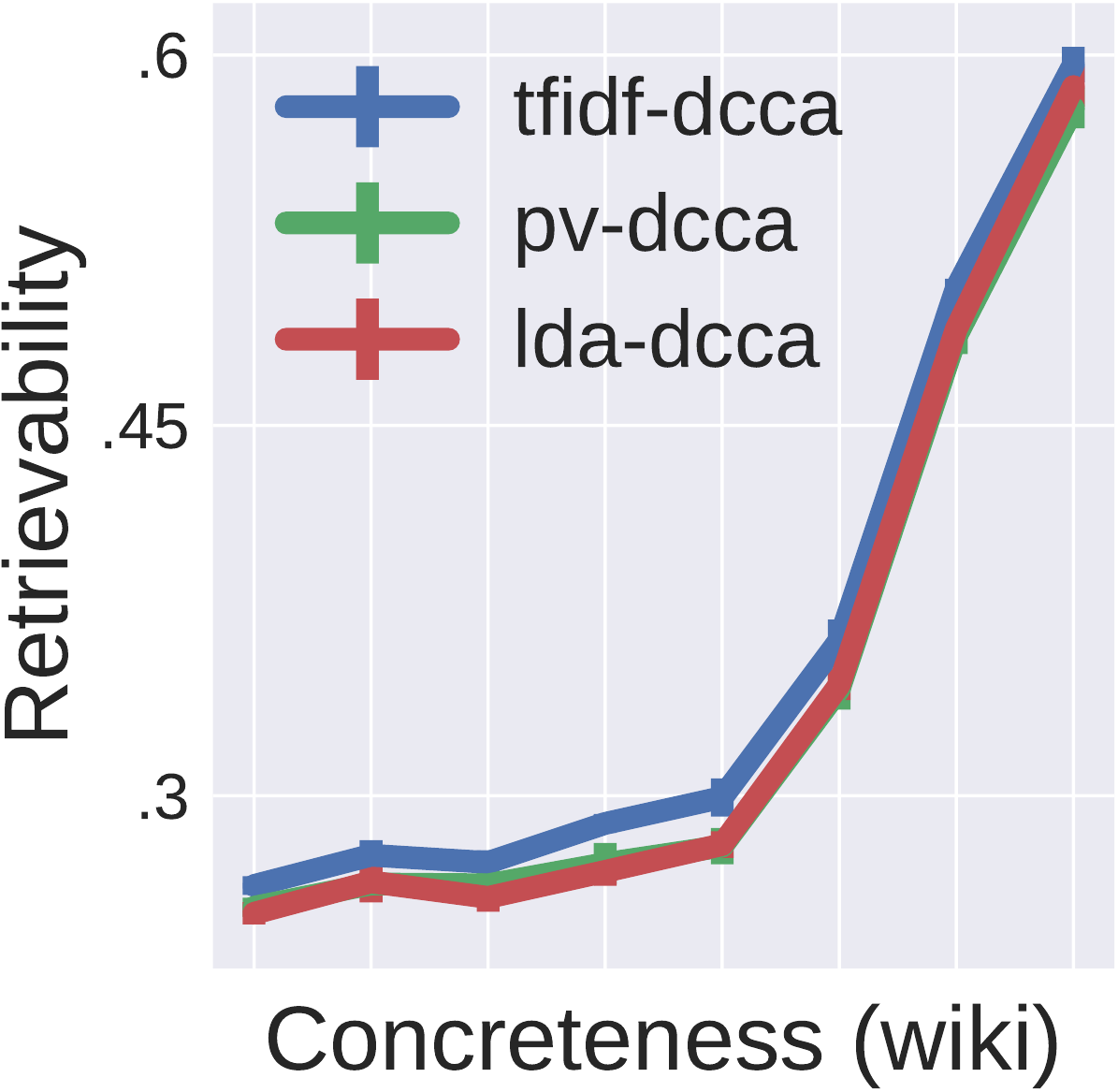}
    \end{minipage}
    \begin{minipage}{0.65\linewidth}
      \centering
      \scriptsize{\begin{tabular}{l|ccccc}
\toprule
 & NP & LS & NS & DCCA\\
\midrule
BTM & 14.1 & 19.8 & 20.7 & 27.9\\
LDA & 19.8 & 36.2 & 33.1 & \thirdplace{37.0}\\
PV & 22.0 & 30.8 & 29.4 & \secondplace{37.1}\\
uni & 17.3 & 29.3 & 30.2 & 36.3\\
tfidf & 18.1 & 35.2 & 33.2 & \firstplace{38.7}\\
\bottomrule
\end{tabular}}
    \end{minipage}
    \caption{Wikipedia}
  \end{subtable}
  \begin{subtable}{.5\linewidth}
    \begin{minipage}{.34\linewidth}
      \centering
      \includegraphics[width=\linewidth]{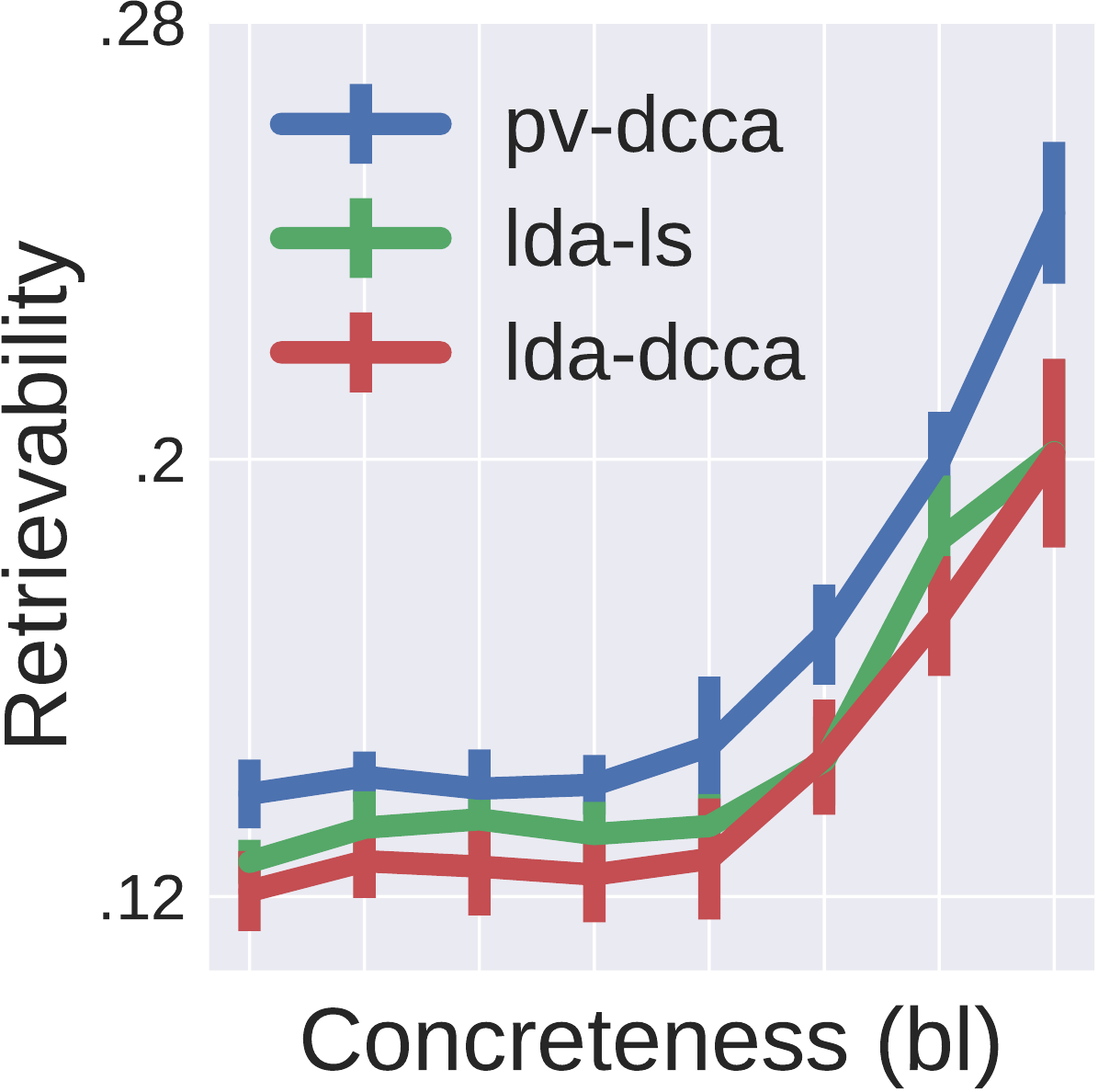}
    \end{minipage}
    \begin{minipage}{0.65\linewidth}
      \centering
      \scriptsize{\begin{tabular}{l|ccccc}
\toprule
 & NP & LS & NS & DCCA\\
\midrule
BTM & 6.7 & 7.3 & 7.2 & 9.5\\
LDA & 10.2 & \secondplace{17.1} & 13.8 & \thirdplace{16.4}\\
PV & 12.6 & 14.1 & 14.1 & \firstplace{17.8}\\
uni & 11.0 & 13.2 & 12.4 & 15.6\\
tfidf & 10.9 & 15.1 & 13.5 & 15.5\\
\bottomrule
\end{tabular}}
    \end{minipage}
    \caption{British Library}
  \end{subtable}
  \\
  \begin{subtable}{.5\linewidth}
    \begin{minipage}{.34\linewidth}
      \centering
      \includegraphics[width=\linewidth]{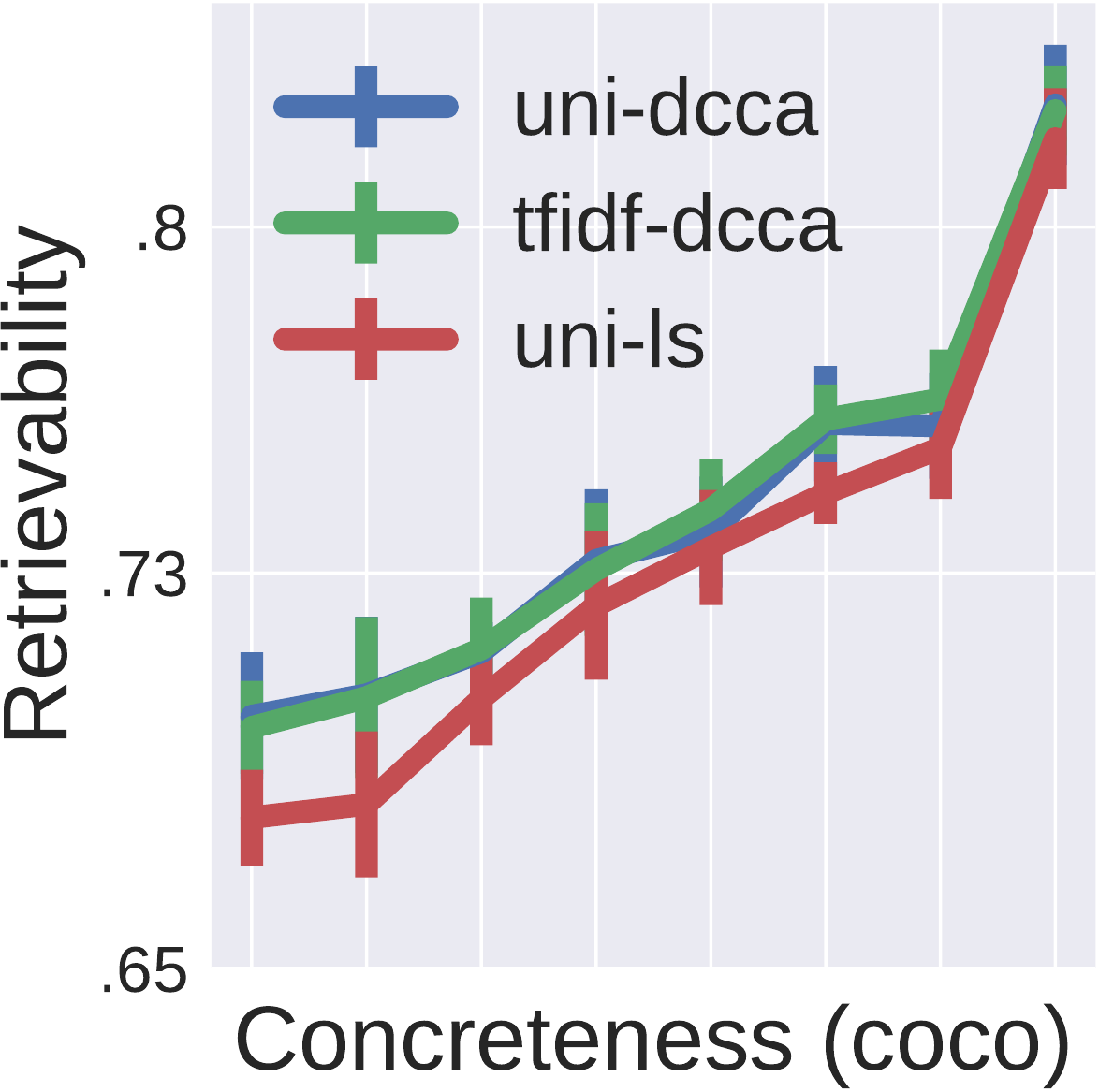}
    \end{minipage}
    \begin{minipage}{0.65\linewidth}
      \centering
      \scriptsize{\begin{tabular}{l|ccccc}
\toprule
 & NP & LS & NS & DCCA\\
\midrule
BTM & 27.3 & 39.9 & 52.5 & 58.6\\
LDA & 23.2 & 51.6 & 51.9 & 51.8\\
PV & 14.1 & 28.4 & 25.7 & 33.5\\
uni & 28.7 & \thirdplace{74.6} & 72.5 & \firstplace{75.0}\\
tfidf & 32.9 & 74.0 & 74.1 & \secondplace{74.9}\\
\bottomrule
\end{tabular}}
    \end{minipage}
    \caption{COCO}
  \end{subtable}
  \begin{subtable}{.5\linewidth}
    \begin{minipage}{.34\linewidth}
      \centering
      \includegraphics[width=\linewidth]{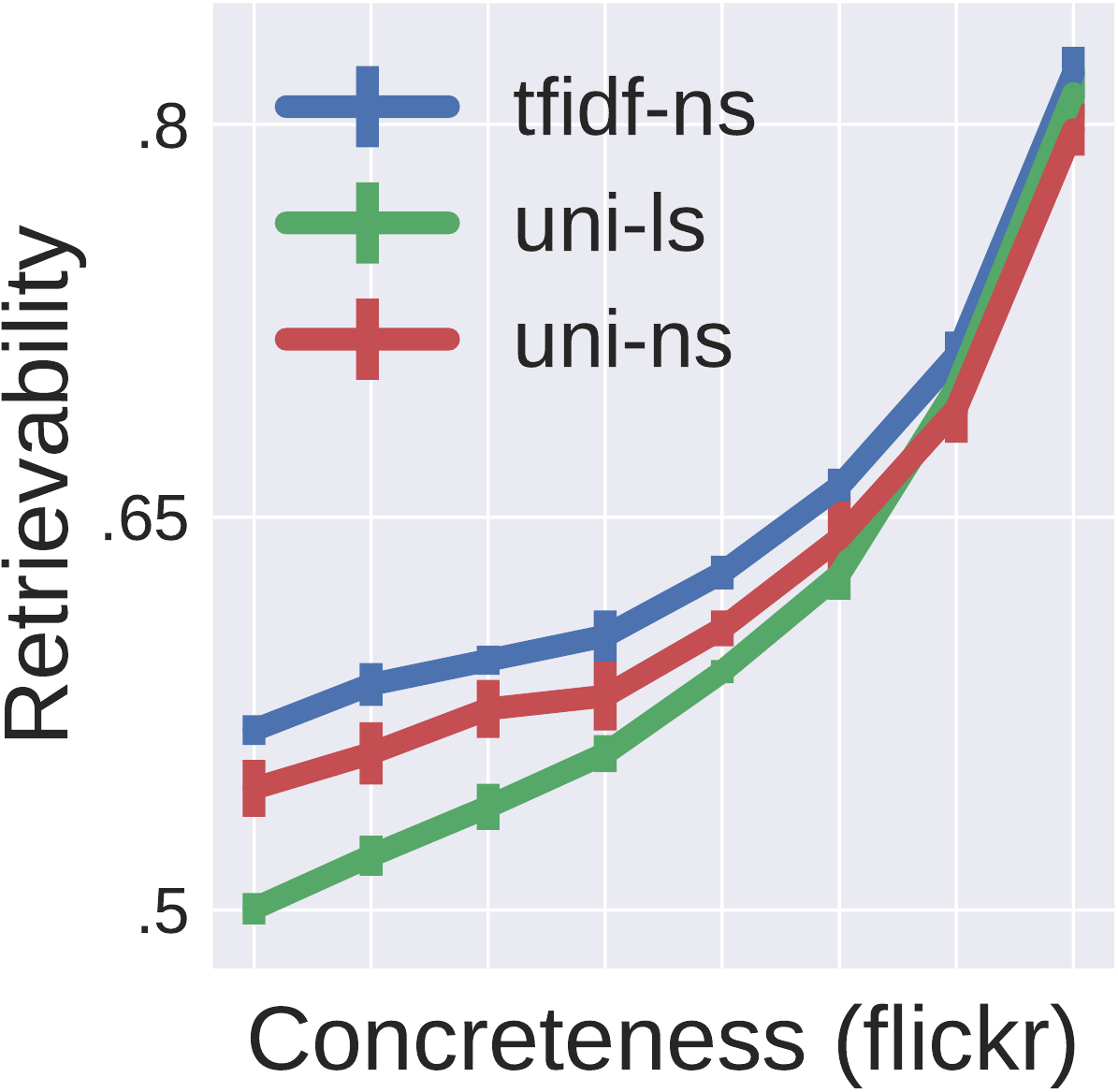}
    \end{minipage}
    \begin{minipage}{0.65\linewidth}
      \centering
      \scriptsize{\begin{tabular}{l|ccccc}
\toprule
 & NP & LS & NS & DCCA\\
\midrule
BTM & 23.9 & 19.1 & 31.0 & 32.4\\
LDA & 18.4 & 32.2 & 34.4 & 34.7\\
PV & 13.9 & 21.3 & 20.0 & 26.6\\
uni & 34.7 & \secondplace{62.5} & \thirdplace{62.0} & 59.6\\
tfidf & 35.1 & 61.6 & \firstplace{63.9} & 60.2\\
\bottomrule
\end{tabular}}
    \end{minipage}
    \caption{Flickr}
  \end{subtable}
  \caption{\Concreteness scores versus \rability (plotted) for each
    dataset, along with Recall at 1\% (in tables, higher is better)
    for each algorithm combination. 
 Tables
    give average retrieval performance over 10-fold cross-validation
    for each combination of NLP/alignment algorithm; the
    \firstplace{best}, \secondplace{second best}, and
    \thirdplace{third best} performing combinations are bolded and
    colored. The \concreteness versus \rability curves are plotted for
    the top-3 performing algorithms, though similar results hold for
    all algorithms. Our \concreteness scores and performance are
    positively correlated, though the shape of the relationship
    between the two differs from dataset to dataset (note the
    differing scales of the y-axes). All results are for RN-ImageNet;
    the similar I3-OpenImages results are omitted for space reasons.}
  \label{fig:bigsweep}
\end{figure*}

\mparagraph{Text Representations}
We consider sparse \textbf{unigram} and \textbf{tfidf} indicator
vectors. In both cases, we limit the vocabulary size to 7.5K. We next
consider latent-variable bag-of-words models, including LDA
\cite{blei2003latent} (256 topics, trained with Mallet
\cite{McCallumMALLET}) a specialized biterm topic model (\textbf{BTM})
\cite{yan2013biterm} for short texts (30 topics), and paragraph
vectors (\textbf{PV}) \cite{le2014distributed} (PV-DBOW version, 256
dimensions, trained with Gensim \cite{rehurek_lrec}).\footnote{We also
  ran experiments encoding text using order-aware recurrent neural
  networks, but we did not observe significant performance
  differences. Those results are omitted for space reasons.}

\mparagraph{Alignment of Text and Images}
We explore four algorithms for learning correspondences between
image and text vectors.
We first compare against \newcite{hodosh2013framing}'s nonparametric
baseline (\textbf{NP}), which is akin to a nearest-neighbor
search. This algorithm is related to the \concreteness score algorithm
we previously introduced in that it exploits the geometry of the
image/text spaces using nearest-neighbor techniques. 
 In general, performance
metrics for this algorithm provide an estimate of how ``easy'' a
particular task is in terms of the initial image/text representations.

We next map image features to text features via a simple linear
transformation.  Let $(t_i, v_i)$ be a text/image pair
in the dataset. We learn a linear transformation $W$ that minimizes
\begin{equation}
  \sum_i \|W\mathfimage(v_i)-\mathftext(t_i)\|_2^2 + \lambda \|W\|_F
  \label{eq:least-squares}
\end{equation}
for feature extraction functions \fimage and \ftext, e.g.,
RN-ImageNet/LDA.
It is possible to map images onto text as in
Equation~\ref{eq:least-squares}, or map text onto images in an
analogous fashion. We find that the directionality of the mapping is
important.
We train models in both directions, and combine their best-performing
results into a single least-squares (\textbf{LS}) model.

Next we consider Negative Sampling (\textbf{NS}), which balances two
objectives: true image/text pairs should be close in the shared latent
space, while randomly combined image/text pairs should be far apart.
For a text/image pair $(t_i, v_i)$, let $s(t_i, v_i)$
be the cosine similarity of the pair in the shared space. The loss for
a single positive example $(t_i, v_i)$ given a
negative sample $(t_i', v_i')$ is
\begin{equation}
h\bigl(s(t_i,v_i), s(t_i,v_i')\bigr) + h\bigl(s(t_i,v_i), s(t_i',v_i)\bigr)
\end{equation}
for the hinge function $h(p,n) = \max\{0, \alpha - p +
n\}$. Following \newcite{kiros2014unifying} we set $\alpha = .2$.

Finally, we consider Canonical Correlation Analysis (CCA), which
projects image and text representations down to independent dimensions
of high multimodal correlation. CCA-based methods are popular within
the IR community for learning multimodal embeddings
\cite{costa2014role,gong2014multi}. We use \newcite{wang2015stochastic}'s
stochastic method for training deep CCA
\cite{andrew2013deep} (\textbf{DCCA}), a method that is competitive
with traditional kernel CCA \cite{wang2015deep} but less
memory-intensive to train.

\mparagraph{Training details} LS, NS, and DCCA were implemented using
Keras \cite{chollet2015keras}.\footnote{%
We used Adam \cite{kingma2014adam},
batch normalization \cite{ioffe2015batch}, and ReLU activations.
Regularization and architectures (e.g., number of layers in DCCA/NS,
regularization parameter in LS) were chosen over a validation set
separately for each cross-validation split. Training is stopped
when retrieval metrics decline over the validation set. All models
were trained twice, using both raw features and
zero-mean/unit-variance features. } In total, we examine all
combinations of: four datasets, five NLP algorithms, two vision
algorithms, four cross-modal alignment algorithms, and two feature
preprocessing settings; each combination was run using 10-fold
cross-validation.


\mparagraph{Absolute retrieval quality} The tables in Figure~\ref{fig:bigsweep}
contain the retrieval results for RN-ImageNet image features across
each dataset, alignment algorithm, and text representation
scheme.
 We show results for $R@1\%$, but $R@5\%$ and $R@10\%$ are
similar. I3-OpenImages image features underperform relative to
RN-ImageNet and are omitted for space reasons, though the results are
similar.

The BL corpus is the most difficult of the datasets we consider,
yielding the lowest retrieval scores. The highly-curated COCO dataset
appears to be the easiest, followed by Flickr and then Wikipedia. No
single algorithm combination is ``best'' in all cases.


\subsection{\Concreteness scores and performance} \label{sec:performance}

We now examine the relationship between retrieval performance and
\concreteness scores. Because \concreteness scores are on the
word/topic level, we define a \emph{\rability} metric that summarizes
an algorithm's performance on a given concept; for example, we might
expect that $\text{\rability}(\text{dog})$ is greater than
$\text{\rability}(\text{beautiful})$.

Borrowing the $R@1\%$ metric from the previous section, we let
$\mathbb{I}[r_i < 1\%]$ be an indicator variable indicating that test
instance $i$ was retrieved correctly, i.e., $\mathbb{I}[r_i < 1\%]$ is
1 if the the average rank $r_i$ of the
image-search-text/text-search-image directions is better than $1\%$,
and 0 otherwise. Let $s_{ic}$ be the affinity of test instance $i$ to
concept $c$. In the case of topic distributions, $s_{ic}$ is the
proportion of topic $c$ in instance $i$; in the case of unigrams,
$s_{ic}$ is the length-normalized count of unigram $c$ on instance
$i$. \Rability is defined using a weighted average over test instances
$i$ as:
\begin{equation}
\text{\rability}(c) = \frac{ \sum_{i} s_{ic} \cdot \mathbb{I}[r_i <
    1\%]}{ \sum_{i} s_{ic} }
\end{equation}
The \rability of $c$ will be higher if instances more associated with
$c$ are more easily retrieved by the algorithm.

\mparagraph{\Rability vs. \Concreteness} The graphs in
Figure~\ref{fig:bigsweep} plot our \concreteness scores versus
\rability of the top 3 performing NLP/alignment algorithm combinations
for all 4 datasets. In all cases, there is a strong positive
correlation between \concreteness and \rability, which provides
evidence that more \concrete concepts are easier to retrieve.

The shape of the \concreteness-\rability curve appears to vary between
datasets more than between algorithms. In COCO, the relationship
between the two appears to smoothly increase. In Wiki, on the other
hand, there appears to be a concreteness threshold, beyond which
retrieval becomes much easier. 

\begin{figure}
  \centering
  \begin{subtable}{.45\linewidth}
    \centering
    \includegraphics[width=.9\linewidth]{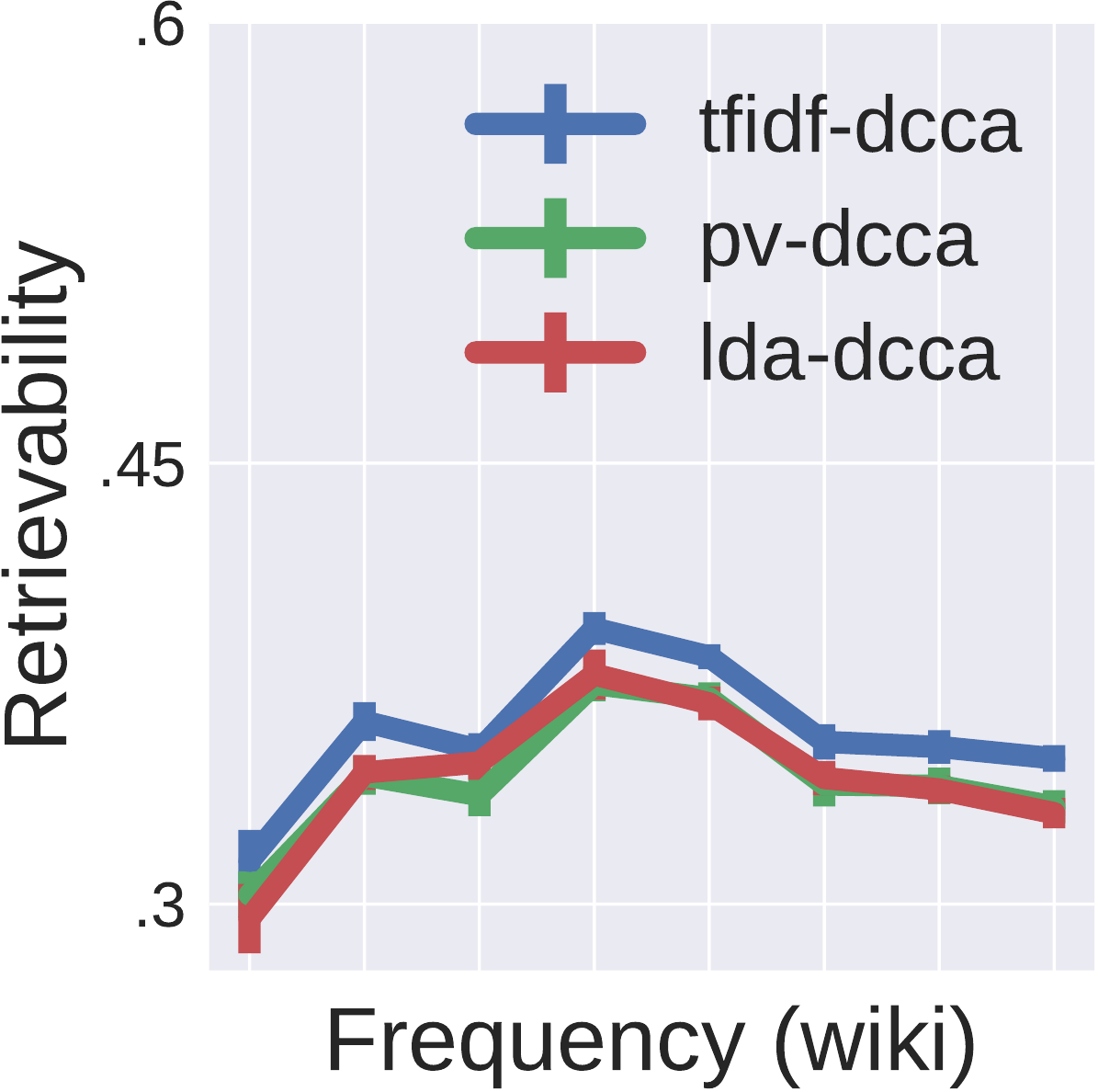}
    \caption{Wikipedia}
  \end{subtable}
  \begin{subtable}{.45\linewidth}
    \centering
    \includegraphics[width=.9\linewidth]{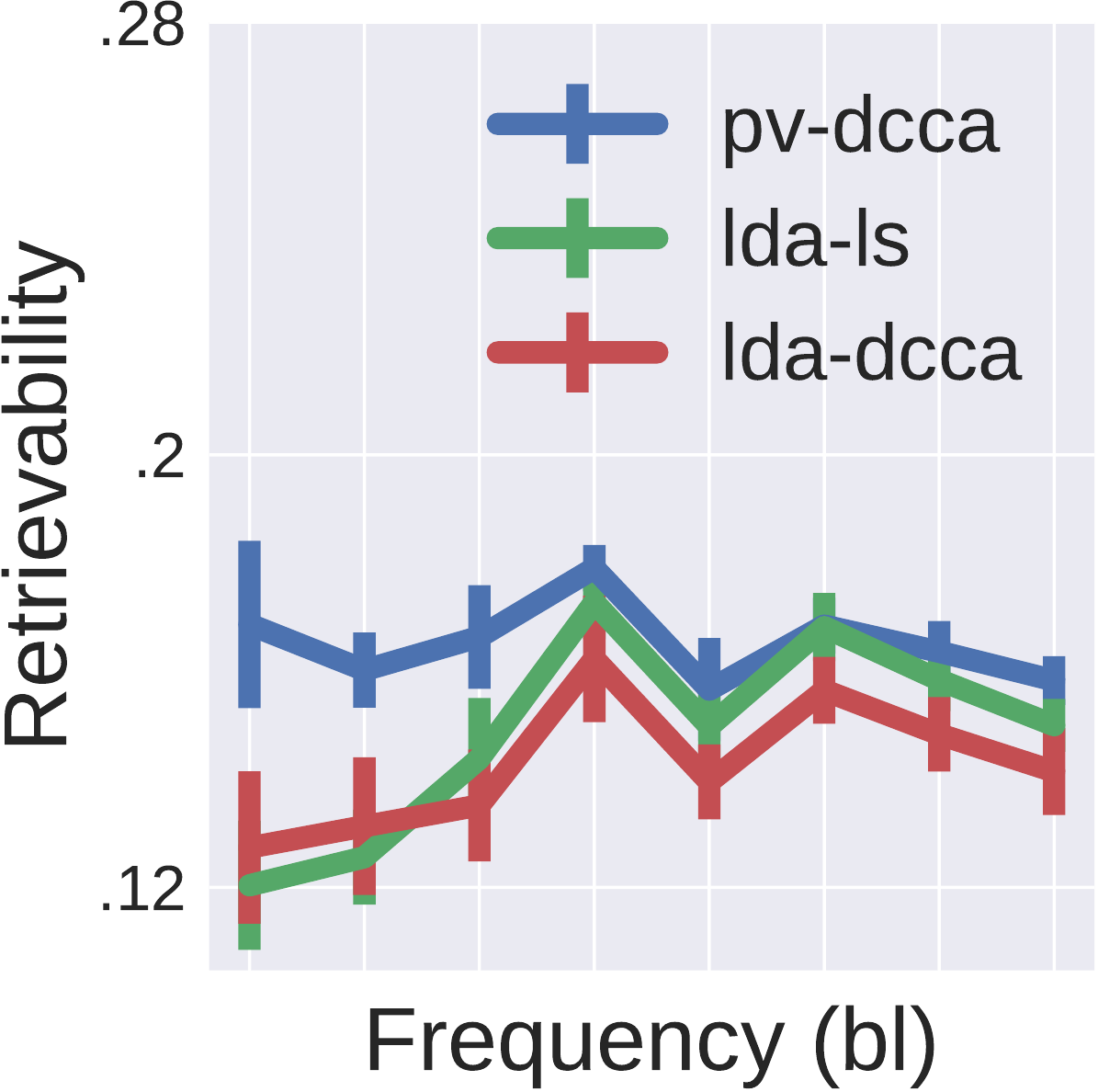}
    \caption{British Library}
  \end{subtable}
  \\
  \begin{subtable}{.45\linewidth}
    \centering
    \includegraphics[width=.9\linewidth]{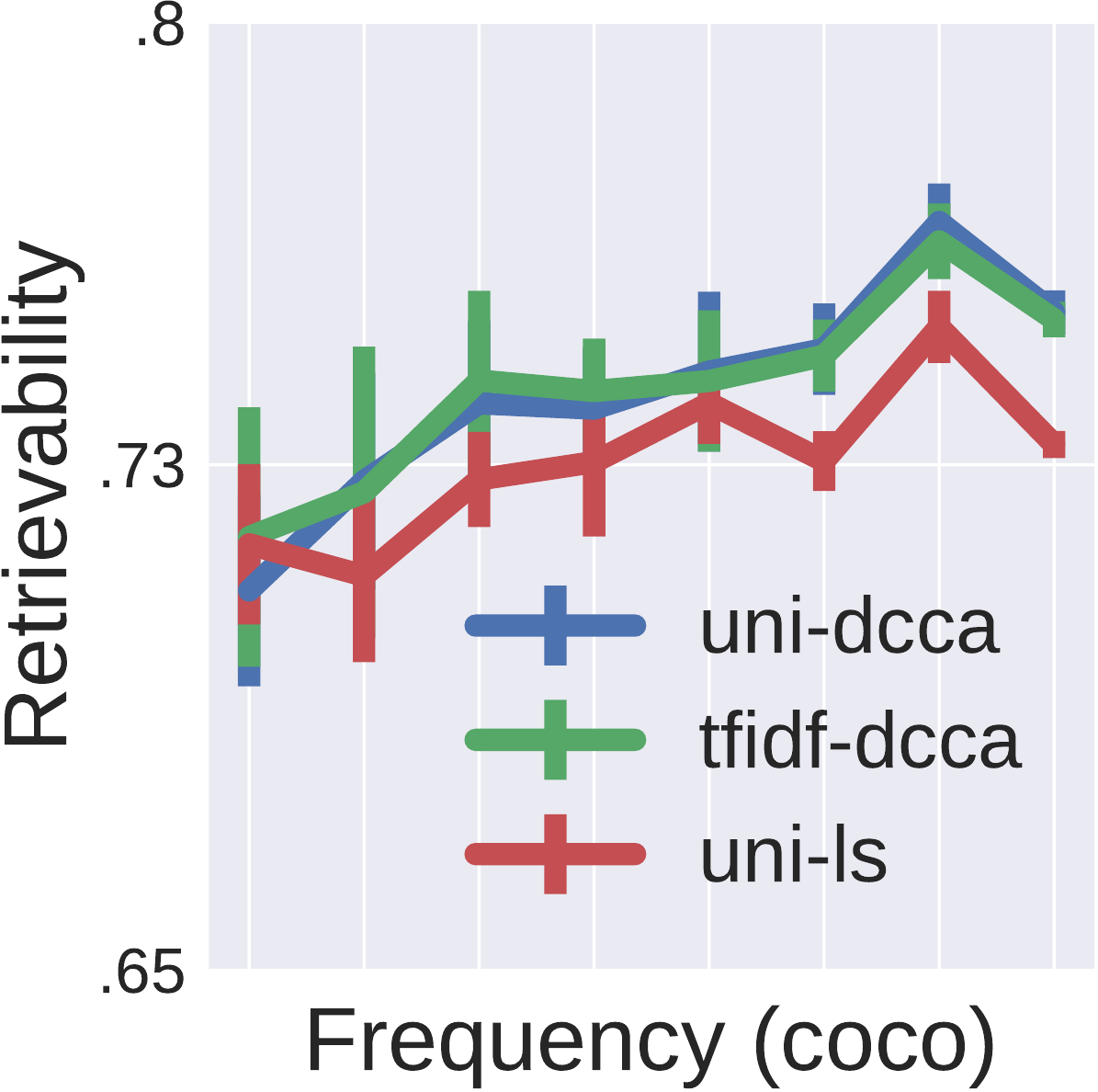}
    \caption{COCO}
  \end{subtable}
  \begin{subtable}{.45\linewidth}
    \centering
    \includegraphics[width=.9\linewidth]{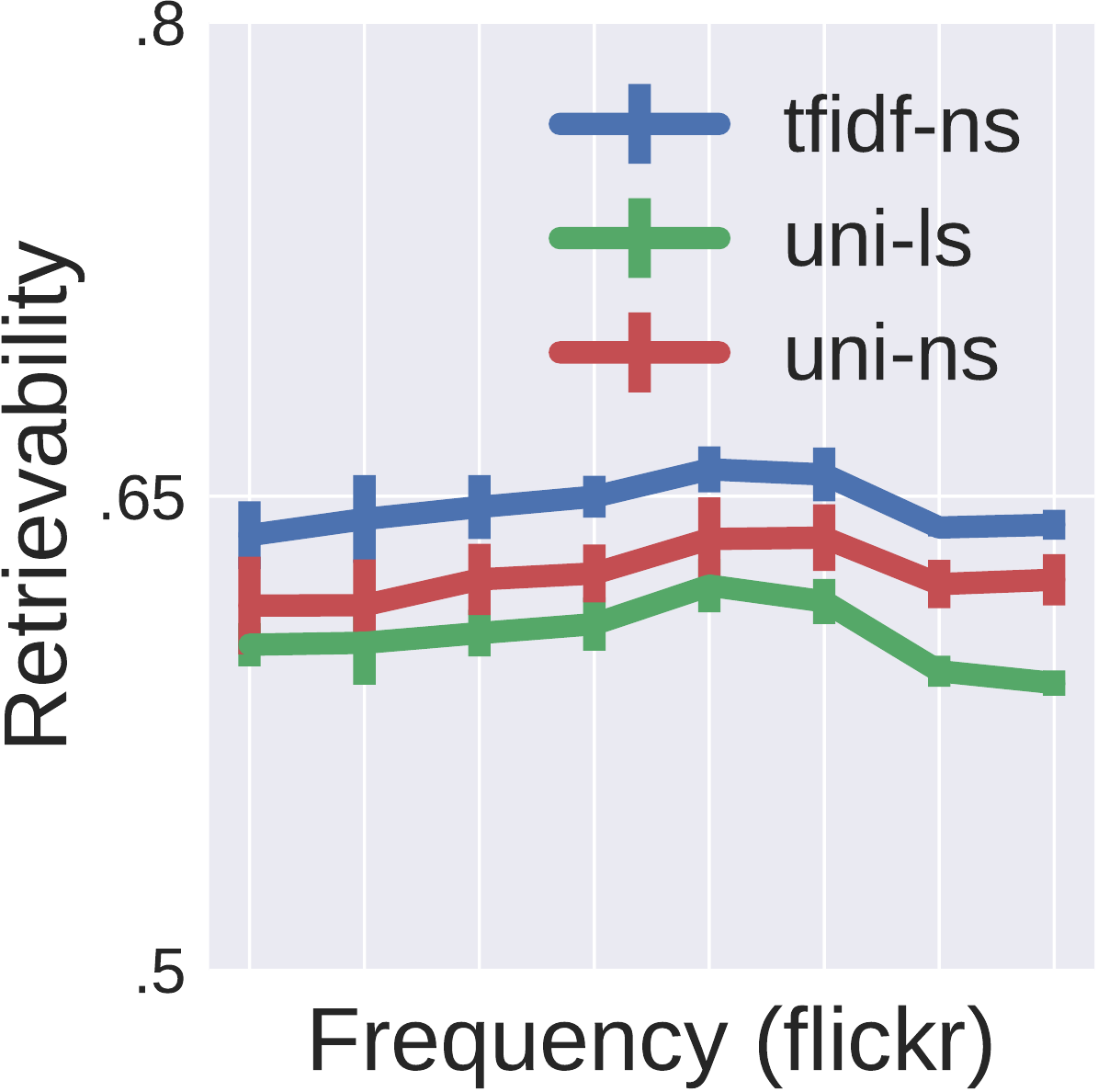}
    \caption{Flickr}
  \end{subtable}
  \caption{Correlation between word/topic frequency and \rability for
    each of the four datasets. Compared to our \concreteness measure
    (see Figure~\ref{fig:bigsweep}; note that the while x-axes are
    different, the y-axes are the same) frequency explains relatively
    little variance in \rability.}
  \label{fig:freq_flatten}
\end{figure}

There is little relationship between \rability and frequency, further
suggesting that our \concreteness measure is not simply mirroring
frequency. We re-made the plots in Figure~\ref{fig:bigsweep}, except
we swapped the x-axis from \concreteness to frequency; the resulting
plots, given in Figure~\ref{fig:freq_flatten}, are much flatter,
indicating that \rability and frequency are mostly uncorrelated.
Additional regression analyses reveal that for the top-3 performing
algorithms on Flickr/Wiki/BL/COCO, concreteness explains
33\%/64\%/11\%/15\% of the variance in \rability, respectively. In
contrast, for all datasets, frequency explained less than 1\% of the
variance in \rability.

\section{Beyond Cross-Modal Retrieval}

Concreteness scores do more than just predict retrieval performance;
they also predict the difficulty of image classification. Two popular
shared tasks from the ImageNet 2015 competition published class-level
errors of all entered systems. We used the unigram \concreteness
scores from Flickr/COCO computed in \S\ref{quantconc} to derive
\concreteness scores for the ImageNet classes.\footnote{There are 1K
  classes in both ImageNet tasks, but we were only able to compute
  \concreteness scores for a subset, due to vocabulary differences.}
We find that for both classification and localization, for all 10 top
performing entries, and for both Flickr/COCO, there exists a
moderate-to-strong Spearman correlation between \concreteness and
performance among the classes for which \concreteness scores were
available ($n_{\text{flickr}},n_{\text{coco}}=171,288$; $.18 < \rho
<.44$; $p<.003$ in all cases). This result suggests that \concrete
concepts may tend to be easier on tasks other than retrieval, as well.


\section{Future Directions}
\label{sec:future}

 At present, it remains unclear if abstract concepts
should be viewed as noise to be discarded (as in Kiela et
al. \shortcite{kiela2014improving}), or more difficult, but learnable,
signal. Because large datasets (e.g., social media) increasingly mix
modalities using ambiguous, abstract language, researchers will need
to tackle this question going forward. We hope that visual
\concreteness scores can guide investigations of the trickiest aspects
of multimodal tasks. Our work suggests the following future
directions:

\mparagraphnp{Evaluating algorithms:} Because \concreteness scores are
able to predict performance prior to training, evaluations could be
reported over concrete and abstract instances separately, as opposed
to aggregating into a single performance metric. A new algorithm that
consistently performs well on non-concrete concepts, even at the
expense of performance on concrete concepts, would represent a
significant advance in multimodal learning.

\mparagraphnp{Designing datasets:} When constructing a new multimodal
dataset, or augmenting an existing one, \concreteness scores can offer
insights regarding how resources should be allocated. Most directly,
these scores enable focusing on ``concrete visual concepts''
\cite{huiskes2010new,chen2015microsoft}, by issuing image-search
queries could be issued exclusively for concrete concepts during
dataset construction. The opposite approach could also be employed, by
prioritizing less concrete concepts.

\mparagraphnp{Curriculum learning:} During training, instances could
be up/down-weighted in the training process in accordance with
\concreteness scores. It is not clear if placing more weight on the
trickier cases (down-weighting concreteness), or giving up on the
harder instances (up-weighting concreteness) would lead to better
performance, or differing algorithm behavior.


\section{Acknowledgments}

This work was supported in part by NSF grants
SES-1741441/IIS-1652536/IIS-1526155, a Yahoo Faculty Research and
Engagement Program grant, the Alfred P. Sloan Foundation, and a
Cornell Library Grant. Any opinions, findings, and conclusions or
recommendations expressed in this material are those of the authors
and do not necessarily reflect the views of the NSF, Sloan Foundation,
Yahoo, or other sponsors. Nvidia kindly provided the Titan X GPU used
in this work.

We would additionally like to thank
Maria Antoniak,
Amit Bahl,
Jeffrey Chen,
Bharath Hariharan,
Arzoo Katiyar,
Vlad Niculae,
Xanda Schofield,
Laure Thompson,
Shaomei Wu,
and the anonymous reviewers
for their constructive comments. 

\bibliographystyle{acl_natbib}
\bibliography{naacl_refs.bib}

\begin{figure*}[b]
\centering
\includegraphics[width=.8\linewidth]{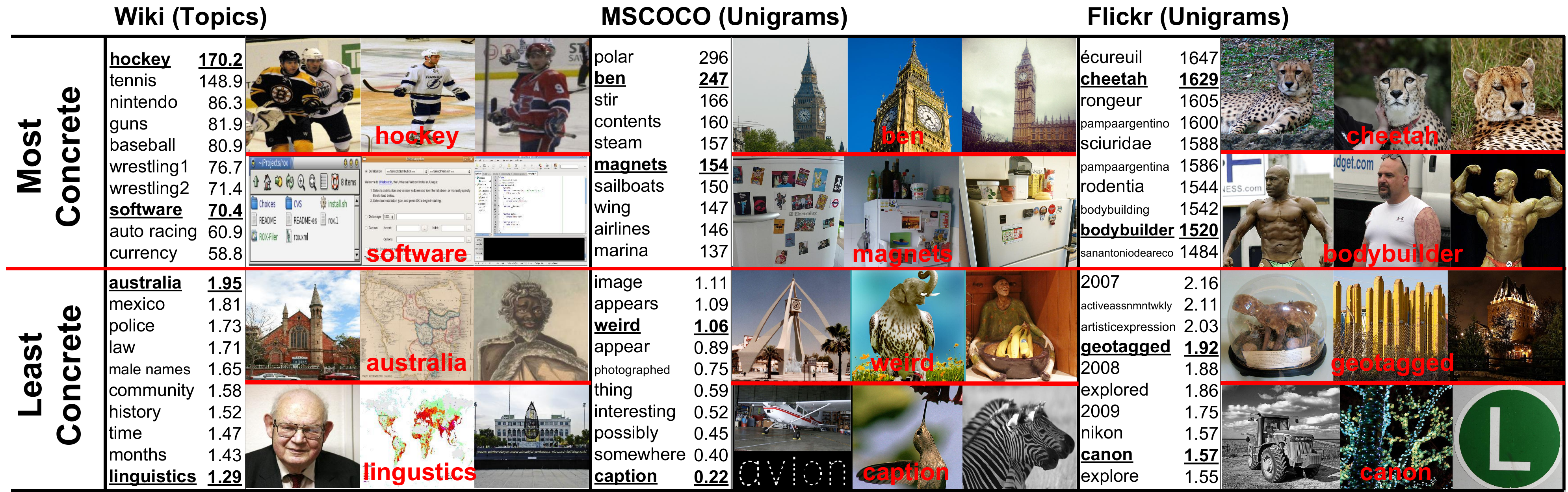}
\captionsetup{labelformat=empty}
\caption{\footnotesize{Images in this figure are reproduced in accordance with the
  following licensing information. The original authors of the images
  do not necessarily endorse this work or the opinions expressed in
  it. Copyright information is given left-to-right. Wikipedia:
  ``hockey" -- Dan4th Nicholas (CC BY 2.0); Leech44 (CC BY 3.0);
  Usasoccer2010 (CC BY-SA 3.0). All ``software" images derive from
  software licensed under GNU GPL v3; screenshot used under fair
  use. ``Australia" -- Toby Hudson (CC BY-SA 3.0); public domain;
  public domain. ``linguistics" -- Suasysar (public domain); USDA
  (public domain); Rama (CC BY-SA 2.0 FR). MSCOCO: ``ben" -- Michael
  Garnett (CC BY-NC-ND 2.0); Giovanni Tufo (CC BY-NC-ND 2.0); Edinho
  Souto (CC BY-NC-SA 2.0). ``magnets" -- Gabriel Pires (CC BY-NC-SA
  2.0); the\_blue (CC BY-NC-SA 2.0); Dennis S. Hurd (CC BY-NC-ND 2.0)
  ``weird" -- waferboard (CC BY 2.0); jaci XIII (CC BY-NC-SA 2.0);
  Ozzy Delaney (CC BY 2.0). ``caption" -- Robert G. Daigle (CC BY 2.0)
  TJ Gehling (CC BY-NC-ND 2.0); amhuxham (CC BY-NC-SA 2.0). Flickr:
  ``cheetah" -- Nathan Rupert (CC BY-NC-ND 2.0); Karen Dorsett (CC
  BY-ND 2.0); Vearl Brown (CC BY-NC 2.0). ``bodybuilder" -- Kent
  (SSCusp) (CC BY-NC-ND 2.0); Frank Carroll (CC BY-NC-SA 2.0); Kent
  (SSCusp) (CC BY-NC-ND 2.0). ``geotagged" -- tpholland (CC BY 2.0);
  s2art (CC BY-NC-SA 2.0); Joe Penniston (CC BY-NC-ND 2.0). ``canon"
  -- ajagendorf25 (CC BY 2.0); Tom Lin (CC BY-NC-ND 2.0); Leo Reynolds
  (CC BY-NC-SA 2.0).}}
\end{figure*}

\begin{figure*}[b]
\centering
\includegraphics[width=.9\linewidth]{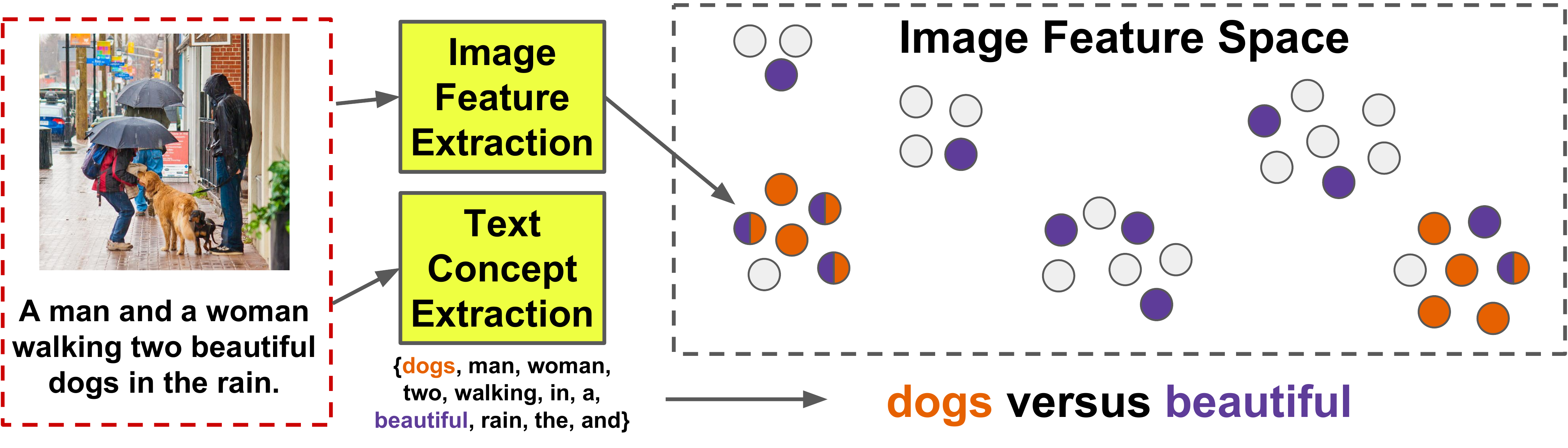}
\captionsetup{labelformat=empty}
\caption{\footnotesize{Images in this figure are reproduced in accordance with the
  following licensing information. The original authors of the images
  do not necessarily endorse this work or the opinions expressed in
  it. J. Michel (aka: Mitch) Carriere (CC BY-NC-ND 2.0)}}
\end{figure*}

\begin{figure*}
\centering
\includegraphics[width=.5\linewidth]{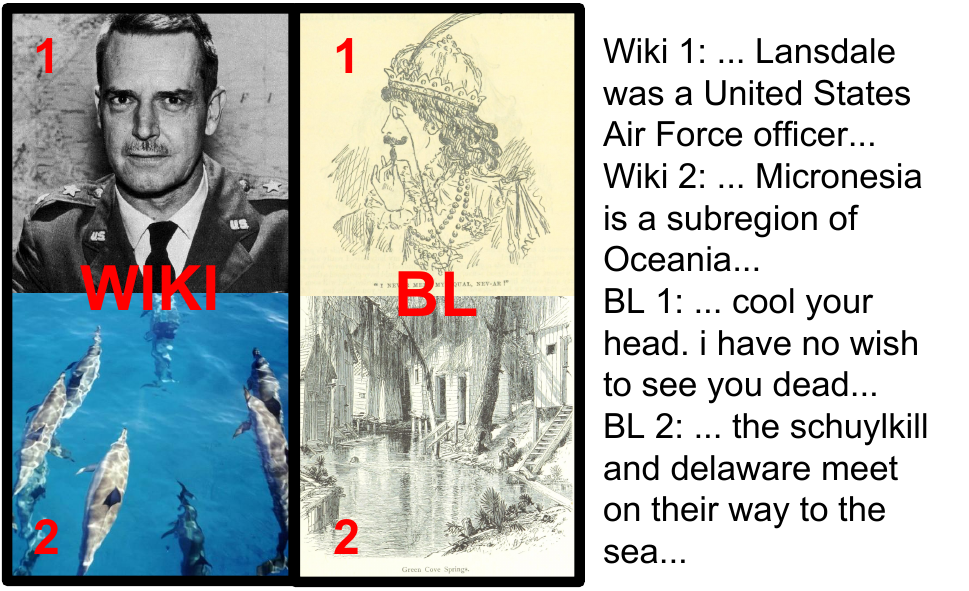}
\captionsetup{labelformat=empty}
\caption{\footnotesize{Images in this figure are reproduced in
    accordance with the following licensing information. The original
    authors of the images do not necessarily endorse this work or the
    opinions expressed in it. Wiki -- US Air Force (public domain);
    Fairsing (public domain). The BL images are also public domain,
    and were originally released by the british library.}}
\end{figure*}

\end{document}